%% file: MAGAR_revised_v2.tex
\documentclass{article}

\usepackage{iclr2027_conference,times}

\input{math_commands.tex}

\usepackage{amsmath}
\usepackage{amssymb}
\usepackage{graphicx}
\usepackage{float}
\usepackage{placeins}
\usepackage{flafter}
\usepackage{booktabs}
\usepackage[table]{xcolor}
\usepackage{multirow}
\usepackage{caption}
\usepackage{url}
\usepackage{makecell}
\usepackage{tcolorbox}
\tcbuselibrary{breakable}

\usepackage{hyperref}

\definecolor{citegreen}{HTML}{507D64}
\definecolor{linkblue}{HTML}{4F7596}

\hypersetup{
  colorlinks=true,
  citecolor=citegreen,
  linkcolor=linkblue,
  urlcolor=linkblue,
  pdfborder={0 0 0}
}

\definecolor{bestcolor}{RGB}{252,239,228}
\definecolor{secondcolor}{RGB}{235,246,229}

\definecolor{promptblue}{RGB}{230,240,255}
\definecolor{promptgray}{RGB}{245,245,245}
\definecolor{outputgreen}{RGB}{235,245,235}

\newtcolorbox{outputbox}{
  colback=outputgreen,
  colframe=outputgreen,
  boxrule=0pt,
  arc=2pt
}

\newtcolorbox{instructionbox}{
  colback=promptblue,
  colframe=promptblue,
  boxrule=0pt,
  arc=2pt
}

\newtcolorbox{inputbox}{
  colback=promptgray,
  colframe=promptgray,
  boxrule=0pt,
  arc=2pt
}

\newtcolorbox{promptbox}[1][]{
  breakable,
  colback=promptgray,
  colframe=black!20,
  boxrule=0.5pt,
  arc=2pt,
  title=#1,
  fonttitle=\bfseries
}

\title{
MAGAR: Multimodal Agents for GUI-Based 3D Mechanical Assembly
and Minimal Rework
}

\author{
\begin{minipage}[t]{\dimexpr\textwidth-2\tabcolsep\relax}
\centering
\bfseries
Jingxuan Wei$^{1,2}$,
Jiahao Fu$^{2}$,
Fuqiang Wang$^{2}$,
Xiaoming Wu$^{2}$,\\
Yuwen Liu$^{2}$,
Kaiwen Wei$^{3}$,
Ming Yang$^{2}$,
Junnan Zhu$^{4}$\thanks{Corresponding author.}\\[6pt]
\normalfont
$^1$ University of Chinese Academy of Sciences, Beijing, China\\
$^2$ Qilu University of Technology (Shandong Academy of Sciences),\\
Shandong Computer Science Center, Jinan, China\\
$^3$ College of Computer Science,
Chongqing University, Chongqing, China\\
$^4$ Institute of Automation,
Chinese Academy of Sciences, Beijing, China\\[4pt]
{\small\texttt{junnan.zhu@nlpr.ia.ac.cn,
weijingxuan20@mails.ucas.ac.cn,
10431250134@stu.qlu.edu.cn}}
\end{minipage}
}

\iclrfinalcopy

\begin{document}

\maketitle

\lhead{Preprint}


\begin{abstract}
A GUI command can be correct at the interface level yet still produce an incorrect assembly. An agent must identify the relevant components, respect their structural relationships, and determine which parts should move during recovery. We introduce \textbf{MAGAR-Bench}, a benchmark of 4,000 multimodal samples from 26 assembly families that separates Assembly Decision, Atomic Action, and Minimal Rework. For rework, the objective counts distinct moved parts within the validated recovery candidate set for each state, rather than the number of actions. We also introduce \textbf{MAGAR}, a selective correction framework built on a frozen policy obtained by supervised fine-tuning. Its key design matches correction granularity to output dependencies. Structured candidates revise assembly fields and rework scope, while complete action candidates keep native GUI primitives coupled with their parameters. A learned selector retains the original prediction unless a candidate passes a task-specific intervention threshold. On recorded states from held-out assembly families, \textbf{MAGAR} improves Overall, the task-balanced aggregate, from 37.50 to 41.39, with gains on all three tasks. Across 209 interventions, 70 correct an error and 22 introduce one. These results support task-specific correction while highlighting the remaining tradeoff between fixing errors and preserving reliable predictions.

\end{abstract}

\section{Introduction}
\label{sec:introduction}

Correct GUI execution does not guarantee correct assembly modification. A command may target a valid interface element while selecting the wrong component, expressing an incompatible assembly intent, or disturbing parts that should remain fixed. When the current assembly is already incorrect, recovery adds another challenge, determining which components should actually move. GUI-based 3D assembly therefore requires more than control recognition. An agent must connect interface operations to object structure, part relationships, and the current assembly state.

Multimodal GUI agents have made substantial progress in visual grounding and computer interaction~\citep{hong2024cogagent,cheng2024seeclick,qin2025uitars}. Benchmarks such as VisualWebArena, OSWorld, and AndroidWorld evaluate interaction in realistic software environments~\citep{koh2024visualwebarena,xie2024osworld,rawles2025androidworld}. Geometry-aware GUI control and assembly error correction further emphasize the importance of structural dependencies~\citep{wei2026pager,wan2024scanet}. Together, these lines of work motivate an evaluation focused on assembly that separates three distinct failures, incorrect assembly intent, incorrect native action grounding, and incorrect recovery scope. Evaluating them separately makes it possible to identify not only whether a prediction improves, but also what was corrected.

\begin{figure}[t]
    \centering
    \includegraphics[width=\linewidth]{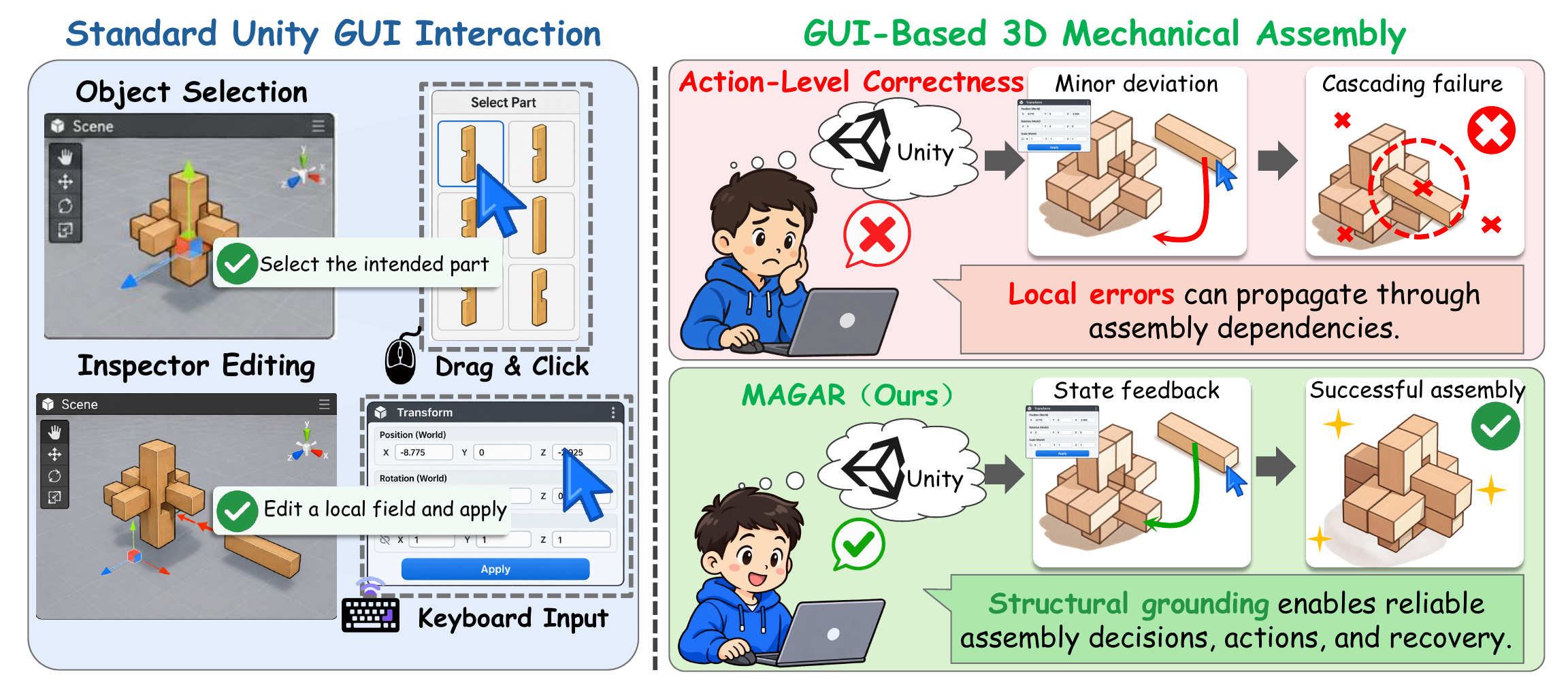}
    \caption{
    Motivation of MAGAR. Locally executable GUI actions may still
    violate the underlying assembly structure.
    }
    \label{fig:motivation}
\end{figure}

We introduce \textbf{MAGAR-Bench} to make these distinctions measurable. Assembly Decision predicts an operation, the involved part set, and the next subgoal; Atomic Action predicts one native GUI primitive and its parameters; Minimal Rework predicts the moved part set and its cardinality, together with auxiliary recovery identifiers. The benchmark contains 4,000 samples from 26 assembly families, with six families held out for testing. Rework is measured by the number of distinct moved parts within the validated recovery candidate set for each state. Moving one part repeatedly is different from moving several parts once. This objective explicitly values preservation of the current assembly. All three tasks are evaluated on recorded Unity states rather than through online rollouts.

The same task decomposition motivates \textbf{MAGAR}, a selective correction framework anchored by a frozen policy obtained through supervised fine-tuning (SFT). Replacing an entire prediction can discard fields that are already correct, while correcting every field independently can violate output dependencies. We therefore align correction granularity with output structure. Module~A proposes structured assembly candidates and field-preserving variants, and revises moved part sets and counts for rework. Module~B proposes complete native actions, keeping each primitive coupled with its parameters. A learned selector retains the original SFT prediction unless a candidate passes a task-specific intervention threshold. Preserving prediction fields is a modeling choice, whereas minimizing moved components is a separate benchmark objective.

Our contributions are summarized as follows.
\begin{itemize}
\item We formulate assembly intent, native GUI action grounding, and minimal rework relative to validated recovery candidates in a multimodal benchmark containing 4,000 samples, with assembly families disjoint across the training and test splits.

\item We develop a selective correction framework that uses structured field corrections for assembly and rework outputs and complete action replacements for native GUI actions, while retaining a common frozen SFT anchor.

\item MAGAR improves the overall task-balanced aggregate, from 37.50 to 41.39 across all three tasks. Intervention accounting further reports both corrected and introduced errors rather than treating every prediction change as beneficial.
\end{itemize}



\section{Related Work}
\label{sec:related_work}

\paragraph{Multimodal GUI Agents.}
Multimodal GUI agents combine visual understanding, language reasoning,
and executable action generation for interaction with digital interfaces.
General-purpose vision-language models, including LLaVA, Qwen2-VL,
Qwen3-VL, DeepSeek-VL2, InternVL2.5, Kimi-VL, MiniCPM-V,
LLaVA-OneVision, and Molmo2, provide the visual representations and
instruction-following capabilities used by such agents
\citep{liu2023llava,bai2024qwen2vl,bai2025qwen3vl,
wu2024deepseekvl2,chen2024internvl25,kimiteam2025kimivl,
yao2024minicpmv,li2024llavaonevision,clark2026molmo2}.
Recent GUI-oriented systems improve screenshot understanding and
interface-element grounding through explicit coordinate prediction or
action-conditioned reasoning, as demonstrated by CogAgent, SeeClick,
and Ferret-UI \citep{hong2024cogagent,cheng2024seeclick,
you2024ferretui}. Agent~S and GUI-Actor further study agentic computer
interaction, while UI-TARS, AgentCPM-GUI, ScaleCUA, and Mobile-Agent-v3.5
extend native GUI control across desktop and mobile environments
\citep{agashe2025agents,wu2025guiactor,qin2025uitars,
zhang2025agentcpmgui,liu2025scalecua,xu2026mobileagent35}.
Related vision-language-action systems, such as OpenVLA and Octo, connect
multimodal observations to executable action policies
\citep{kim2024openvla,team2024octo}. These methods primarily target
general action grounding and task completion. MAGAR instead separates
structured assembly intent, atomic GUI execution, and minimal rework
prediction, allowing intervention modules to correct different classes
of errors while preserving reliable SFT outputs.

\paragraph{Benchmarks.}
Existing benchmarks evaluate multimodal agents across web, desktop,
mobile, and general assistant settings. WebArena, VisualWebArena, and
Mind2Web focus on browser-based task completion and web interaction
\citep{zhou2024webarena,koh2024visualwebarena,deng2023mind2web}.
OSWorld and AndroidWorld provide execution-based evaluation for desktop
and mobile environments, respectively, while BrowserGym studies
interactive browser tasks \citep{xie2024osworld,rawles2025androidworld,
si2024browsergym}. GAIA evaluates broader assistant capabilities that
require multimodal reasoning and tool use, and AgentBench measures
language-model agents across multiple environments
\citep{mialon2024gaia,liu2024agentbench}. Assembly-oriented benchmarks
introduce additional structural requirements: SCANet studies error
identification and correction in LEGO assembly, while FurnitureBench
evaluates long-horizon physical furniture assembly
\citep{wan2024scanet,heo2025furniturebench}. These benchmarks measure
functional completion, grounding, or physical assembly success, but do
not jointly evaluate structured assembly decisions, native GUI action
parameters, and the number of distinct parts requiring recovery.
MAGAR-Bench addresses this gap through an assembly-family-disjoint
evaluation setting with separate targets for Assembly Decision, Atomic
Action, and Minimal Rework.


\section{Method}
\label{sec:method}

\subsection{Problem Formulation}
\label{sec:preliminaries}

We study task-specific prediction for GUI-based 3D mechanical assembly.
Each example contains the current assembly state, the target assembly
state, candidate-part references, public GUI information, and summarized
interaction history. We denote the multimodal input and task output by
\begin{equation}
\begin{aligned}
X&=(I_c,I_g,I_p,S,H),
\qquad
I_p=\{I_p^j\}_{j=1}^{N_p},\\
t&\in
\{
\mathrm{AssemblyDecision},
\mathrm{AtomicAction},
\mathrm{MinimalRework}
\},\\
y_t&=
\begin{cases}
(o,P,g), & t=\mathrm{AssemblyDecision},\\
(a,\theta), & t=\mathrm{AtomicAction},\\
(P_m,n_m,R,A), & t=\mathrm{MinimalRework}.
\end{cases}
\end{aligned}
\label{eq:method_input}
\end{equation}
Here, $o$ is the assembly operation, $P$ is the involved-part set,
and $g$ is the next subgoal. For Atomic Action, $a$ is a GUI
primitive and $\theta$ contains its typed parameters. For Minimal Rework, $P_m$ denotes the moved-part set and $n_m$ denotes
the minimum number of distinct moved parts required for recovery among
the validated recovery candidates for the state. The
output also includes a recovery route $R$ and corrective action sequence
$A$ to describe the associated recovery process. This task measures
whether an agent can identify the necessary components to move while
preserving the existing assembly as much as possible.
The rework objective counts distinct moved parts rather than the total
number of actions. Let $\mathcal{R}_{\mathrm{val}}(X)$ denote the
validated recovery candidates associated with state $X$. For a validated
recovery candidate
$\pi=(a_1,\ldots,a_{L_\pi})$, we define the recovery cost as

\begin{equation}
\begin{aligned}
J_{\mathrm{move}}(\pi)
&=
\left|
\bigcup_{\ell=1}^{L_\pi}
P_{\mathrm{move}}(a_\ell)
\right|,
\\[2mm]
\pi^{*}
&=
\underset{\pi\in\mathcal{R}_{\mathrm{val}}(X)}{\arg\min}
\;J_{\mathrm{move}}(\pi).
\end{aligned}
\label{eq:method_anchor_objective}
\end{equation}
\begin{figure}[t]
    \centering
    \includegraphics[width=\linewidth]{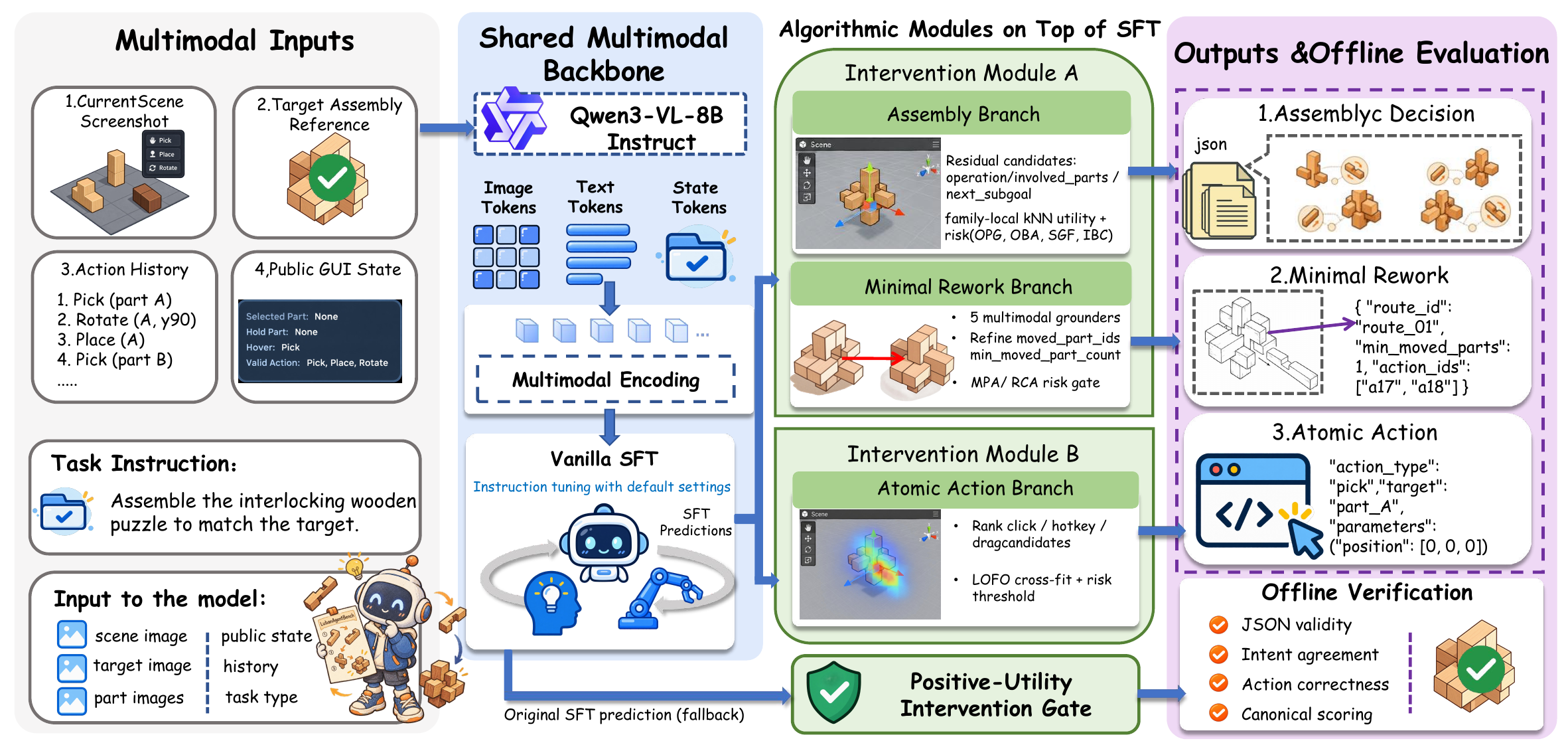}
    \caption{
Overview of MAGAR. A frozen SFT backbone provides task-specific
predictions, while Module~A performs structured assembly and rework
intervention and Module~B performs complete-action intervention.
    }
    \label{fig:method_overview}
\end{figure}

\subsection{MAGAR Framework}
\label{sec:magar_framework}

MAGAR is a two-module intervention framework built on a supervised
fine-tuned \texttt{Qwen3-VL-8B-Instruct} backbone
\citep{bai2025qwen3vl}. Given a multimodal input $X$ and a task family
$t$, the frozen backbone first produces an anchor prediction
$y_{\mathrm{SFT}}=f_{\theta_0}(X,t)$. MAGAR then constructs a
task-specific candidate set
$\mathcal{C}_t(X)=\operatorname{Cand}_t(X,y_{\mathrm{SFT}})$
around this anchor and estimates whether replacing the anchor is
likely to improve the corresponding task metrics. The candidate with
the highest estimated utility is accepted only when it satisfies the
task-specific reliability gate:
\begin{equation}
\widehat{y}
=
\begin{cases}
c^{*},
&
\Gamma_t(c^{*}\mid X,y_{\mathrm{SFT}})=1,\\
y_{\mathrm{SFT}},
&
\text{otherwise},
\end{cases}
\qquad
c^{*}
=
\arg\max_{c\in\mathcal{C}_t(X)}
\overline{U}_t(c\mid X,y_{\mathrm{SFT}}).
\label{eq:magar_selective_prediction}
\end{equation}

The candidate generator is task-specific. Module~A handles structured
Assembly Decision and Minimal Rework outputs, whereas Module~B handles
Atomic Action outputs. Both modules use the same SFT prediction as the
preservation anchor, making intervention optional rather than
mandatory. The methodological contribution is the task-specific design
of these two modules; the utility gate is a common rule for deciding
when a generated candidate should replace the anchor.

For each task, the selector predicts a vector of metric changes
$\widehat{\boldsymbol{\Delta}}_t$ and a benefit probability
$\widehat{p}^{+}_t$. The utility and reliability gate are defined as
\begin{equation}
\begin{aligned}
\overline{U}_t(c\mid X,y_{\mathrm{SFT}})
&=
\mathbf{w}_t^{\top}
\widehat{\boldsymbol{\Delta}}_t
(c\mid X,y_{\mathrm{SFT}}),\\
\Gamma_t(c\mid X,y_{\mathrm{SFT}})
&=
\mathbb{I}\!\left[
\overline{U}_t(c\mid X,y_{\mathrm{SFT}})\geq\tau_t,\;
\widehat{p}^{+}_t(c\mid X,y_{\mathrm{SFT}})\geq\tau_t^{+}
\right],
\end{aligned}
\label{eq:magar_utility_gate}
\end{equation}
where $\mathbf{w}_t$ contains the task-specific metric weights,
$\tau_t$ is the minimum predicted utility, and $\tau_t^{+}$ is the
minimum benefit probability required for intervention.

The selector supervision is constructed from candidates generated for
training examples. Let $\mathcal{M}_t$ denote the metrics associated
with task $t$, $S_m$ the score under metric $m$, and $y_i^{*}$ the
reference target of training example $i$. The realized metric-change
vector and benefit label are
\begin{equation}
\begin{aligned}
\boldsymbol{\Delta}_i(c)
&=
\left(
S_m(c,y_i^{*})
-
S_m(y_{\mathrm{SFT},i},y_i^{*})
\right)_{m\in\mathcal{M}_{t_i}},\\
b_i(c)
&=
\mathbb{I}\!\left[
\mathbf{w}_{t_i}^{\top}
\boldsymbol{\Delta}_i(c)>0
\right].
\end{aligned}
\label{eq:selector_targets}
\end{equation}
The metric-change head is trained by regression, while the reliability
head predicts whether the candidate improves the weighted task score:
\begin{equation}
\mathcal{L}_{\mathrm{sel}}
=
\sum_{i}
\sum_{c\in\mathcal{C}_{t_i}(X_i)}
\left[
\left\|
\widehat{\boldsymbol{\Delta}}_{t_i}(c)
-
\boldsymbol{\Delta}_i(c)
\right\|_2^2
+
\lambda_{\mathrm{rel}}
\operatorname{BCE}
\left(
\widehat{p}^{+}_{t_i}(c),b_i(c)
\right)
\right].
\label{eq:selector_training}
\end{equation}
During selector training, candidate-level targets and the selector loss are
constructed from training-family examples under family-level cross-fitting.
After calibration, the selector and gate thresholds are fixed before
evaluation.

The gate makes preservation an explicit option. If the highest-utility
candidate fails either reliability condition, MAGAR returns the
original SFT output. This design separates candidate coverage from
selection reliability: the intervention modules propose possible
corrections, while the selector determines whether a proposed change
has sufficient evidence to replace the anchor.

MAGAR reuses hidden representations from the frozen SFT backbone
instead of performing a second autoregressive generation step. The
hidden-token spans of the current state, target state, public GUI
context, and candidate-part references are pooled into $h_c$, $h_g$,
$h_p$, and $\{h_r^j\}_{j=1}^{N_p}$, respectively. Multi-head attention
aligns the current and target states to produce their residual
difference $d$, while attention pooling aggregates the reference-part
features into $\bar{h}_r$. The task-conditioned plan is constructed as
\begin{equation}
\begin{aligned}
z_s &= \phi_z\left([h_c;h_g;h_p;\bar{h}_r;d;e_t]\right),\\
z_A &= z_s+\rho_t(z_s).
\end{aligned}
\label{eq:frozen_task_representation}
\end{equation}
where $e_t$ is the task-family embedding, $z_s$ is the shared
multimodal plan, and $\rho_t$ is a lightweight task-specific residual
adapter initialized with zero impact. The resulting representation
$z_A$ summarizes the current-to-target change, candidate-part evidence,
GUI context, and task identity while preserving compatibility with the
frozen SFT representation at the start of training. Module~A uses
$z_A$ for structured assembly and rework prediction, while Module~B
uses it to condition atomic-action grounding.
\subsection{Module A: Structured Assembly and Rework Intervention}
\label{sec:module_a}

Module~A addresses the two outputs that require explicit structural
reasoning: Assembly Decision and Minimal Rework. For Assembly Decision,
the module predicts an operation, a compatible next subgoal, and an
operation-conditioned set of involved parts. These fields are scored as
one structured object rather than independently concatenated after
decoding.
\begin{equation}
p_A(o,P,g\mid X)
\propto
p_o(o\mid z_A)
p_g(g\mid o,z_A)
p_P(P\mid o,X,z_A)
\mathbb{I}[(o,g)\in\mathcal{R}_{\mathrm{valid}}].
\label{eq:module_a_assembly_distribution}
\end{equation}
where $\mathcal{R}_{\mathrm{valid}}$ is the set of valid
operation--subgoal pairs. The operation-conditioned part scores use the
reference-part representations together with the current--target
difference, and $p_K$ predicts the cardinality of the selected set.
The resulting candidates are not required to replace the complete SFT
answer. For each candidate, Module~A enumerates residual variants that
replace a non-empty subset of the fields $\{o,P,g\}$ while retaining
all other anchor fields. This allows the module to correct an incorrect
operation without discarding a correct part set, or to correct the part
set while preserving the anchor subgoal.

For Minimal Rework, Module~A proposes moved-part sets and their counts.
Five independently trained part grounders generate candidates
from reference-part evidence and current--target differences. To retain
the full output format, each candidate inherits the recovery route and
corrective action sequence from the SFT anchor. Let
$\mathcal{K}_q(X,z_A)$ denote the top-ranked moved-part and count pairs
from the $q$-th Rework grounder. The candidate set is
\begin{equation}
\mathcal{C}_{A}^{\mathrm{rew}}(X)
=
\bigcup_{q=1}^{5}
\left\{
(P_m,n_m,R_{\mathrm{SFT}},A_{\mathrm{SFT}})
\mid
(P_m,n_m)\in\mathcal{K}_q(X,z_A)
\right\}.
\label{eq:module_a_rework_candidates}
\end{equation}
Assembly and Rework candidates are passed to task-specific selectors,
which estimate gains in the corresponding evaluated metrics and apply
the utility gate in Eq.~\ref{eq:magar_utility_gate}. For Minimal Rework, Module~A predicts the recovery scope through the
moved-part set and its corresponding count. The recovery route and
corrective action sequence are inherited from the SFT anchor and kept
as auxiliary recovery information. Thus, Module~A improves component-level
rework identification while preserving the original recovery trajectory.
\subsection{Module B: Plan-Conditioned Atomic Action Intervention}
\label{sec:module_b}

Module~B handles Atomic Action prediction. The action space contains
\texttt{click}, \texttt{drag}, \texttt{type\_text}, \texttt{keypress}, and
\texttt{hotkey}. Each action is represented as a complete primitive--
parameter pair rather than as independently selected fields. The action
router first selects a primitive, after which the corresponding expert
predicts its valid parameters. We define the action state as
$h_B=\phi_B[h_c;h_p;z_A;\phi_S(S);\phi_H(H)]$. The resulting action
distribution is
\begin{equation}
p_B(a,\theta\mid X)
=
p_{\mathrm{router}}(a\mid h_B)\,
p_{\mathrm{expert}}(\theta\mid a,h_B,z_A).
\label{eq:module_b_action_policy}
\end{equation}
The plan representation $z_A$ conditions both primitive selection and
parameter prediction. For each action group $k$, Module~B introduces a
small gated residual into the corresponding logits:
\begin{equation}
\ell_k'
=
\ell_k+
\tanh(\alpha_k)W_kz_A,
\qquad
k\in
\{
\mathrm{router},
\mathrm{click},
\mathrm{drag},
\mathrm{type},
\mathrm{keypress},
\mathrm{hotkey}
\}.
\label{eq:module_b_plan_conditioning}
\end{equation}
The gates and residual parameters are initialized near zero, allowing
the action module to preserve the behavior learned by SFT before
learning to exploit the structural plan. During training, Module~A is
frozen while Module~B learns to consume its plan representation.
At inference, Module~B ranks complete action candidates with a
task-specific selector and replaces the SFT action only when the
predicted improvement satisfies the corresponding risk constraints.
\section{Dataset}
\label{sec:dataset}

\subsection{Dataset Construction}
\label{sec:dataset_construction}

\begin{figure}[ht]
    \centering
    \includegraphics[width=\linewidth]{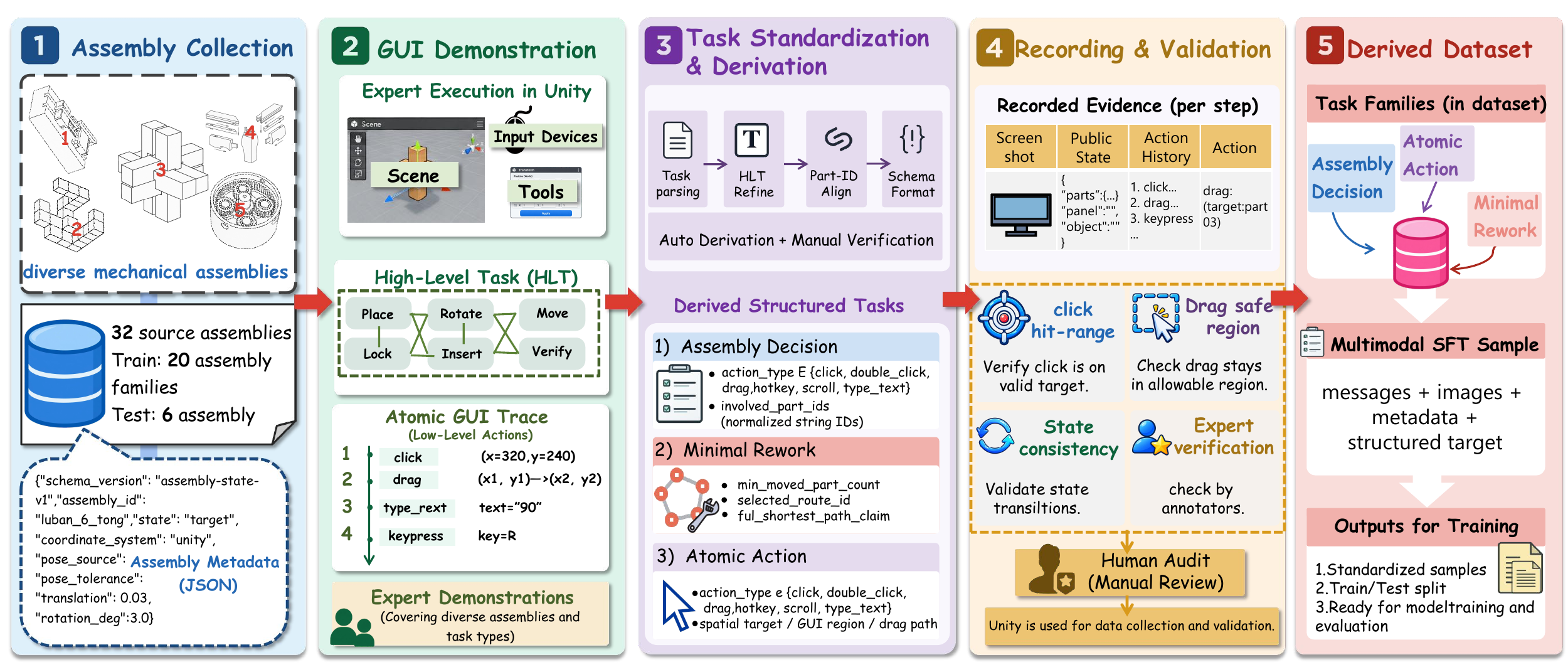}
    \caption{Construction pipeline of MAGAR-Bench.}
    \label{fig:dataset_construction}
\end{figure}

MAGAR-Bench is constructed from Unity-based 3D mechanical assembly
episodes that record interaction with virtual components through
native GUI operations. Figure~\ref{fig:dataset_construction}
summarizes the construction pipeline. The dataset links assembly
observations and public state to decision, action, and recovery
targets, preserving the context needed to interpret each prediction.
\noindent\textbf{Episode recording and contextual organization.}
Each episode records the target assembly, intermediate states,
interaction trajectories, and GUI actions executed in Unity.
Task-specific inputs retain the relevant observations, part
references, public state, and history.
\noindent\textbf{Assembly-decision targets.}
Assembly Decision pairs a recorded state with the next structured
intent: the operation, involved-part set, and next subgoal.
These targets retain the association between an intended operation
and the components to which it applies.

\noindent\textbf{Atomic-action targets.}
Atomic Action pairs an interface state and assembly objective with
a native operation and its parameters. The action grammar preserves
the parameterization of spatial, keyboard, and text interactions.

\noindent\textbf{Minimal-rework targets.}
Minimal Rework pairs an incorrect assembly state with a moved-part
set and the minimum number of distinct parts that must move among the
validated recovery candidates associated with that state.
Recorded recovery routes and corrective actions are retained as output
fields and are scored only through the identifier-matching terms in RCA;
their execution validity is not evaluated. The three
task families provide separate prediction-level measurements within
the same assembly domain.
\subsection{Dataset Analysis}
\label{sec:dataset_analysis}
\vspace{-5mm}

\noindent
\begin{minipage}[t]{0.49\columnwidth}
\vspace{0pt}
Figure~\ref{fig:dataset_analysis} summarizes
\textsc{MAGAR-Bench}. The benchmark contains 4,000 samples from 26
assembly families, with 3,000 samples from 20 families for training
and 1,000 samples from six held-out families for testing. It includes
974 Assembly Decision, 2,426 Atomic Action, and 600 Minimal Rework
samples, covering high-level decisions, low-level execution, and
assembly recovery. Atomic Action covers five native GUI operations:
\texttt{click}, \texttt{drag}, \texttt{hotkey}, \texttt{keypress}, and
\texttt{type\_text}. Minimal Rework includes recovery cases involving
different numbers of moved parts.
\end{minipage}
\hfill
\begin{minipage}[t]{0.49\columnwidth}
\vspace{0pt}
\centering
\includegraphics[width=\linewidth]{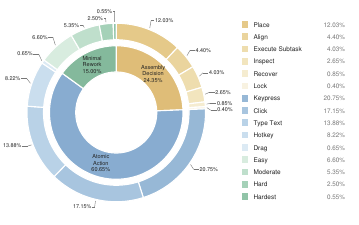}
\captionof{figure}{Composition of MAGAR-Bench.}
\label{fig:dataset_analysis}
\end{minipage}

\FloatBarrier

\section{Experiments}
\label{sec:experiments}

\subsection{Experimental Setup}
\label{sec:experimental_setup}
We evaluate Assembly Decision, Atomic Action, and Minimal Rework on
recorded task instances from held-out assembly families. MAGAR is built
on a supervised fine-tuned \texttt{Qwen3-VL-8B-Instruct}
backbone~\citep{bai2025qwen3vl}. All intervention configurations use
the same frozen SFT anchor and activate the applicable module for each
task. The task-specific selectors are trained through cross-fitting on
the training families. Open-source multimodal models, proprietary
models, GUI-specialized baselines, and MAGAR follow the same held-out
test protocol, while LLM-as-judge evaluation uses a separate fixed
subset. Overall is computed by first averaging the automatic metrics
within each task family and then assigning equal weight to the three
task-family scores; the LLM-as-judge score is reported separately.
Implementation details and metric definitions are provided in the
appendix.

\subsection{Main Results}
\label{sec:main_results}

Table~\ref{tab:main_results} reports the main results. MAGAR
reaches 41.39 Overall, improving on Vanilla-SFT's 37.50 by 3.89 points
under the reported cross-fitted protocol. It has the highest reported
Overall among the systems evaluated on the complete Test split.
The improvement is distributed across the three task families.

\begin{table*}[!htbp]
\centering
\footnotesize
\caption{
Main results on recorded task instances. Models are grouped into open-source
multimodal models, proprietary multimodal models, and GUI agent baselines.
The best results are highlighted in bold, and the second-best results are
underlined.
}
\label{tab:main_results}
\setlength{\tabcolsep}{2.5pt}
\renewcommand{\arraystretch}{0.95}
\resizebox{\textwidth}{!}{
\begin{tabular}{l|cccc|ccc|cc|c|c}
\toprule
\multirow{2}{*}{\textbf{Model}}
& \multicolumn{4}{c|}{\textbf{Assembly Decision}}
& \multicolumn{3}{c|}{\textbf{Atomic Action}}
& \multicolumn{2}{c|}{\textbf{Minimal Rework}}
& \multirow{2}{*}{\textbf{Overall}$\uparrow$}
& \textbf{LLM-as-judge} \\
\cmidrule(lr){2-5}
\cmidrule(lr){6-8}
\cmidrule(lr){9-10}
& OPG$\uparrow$
& OBA$\uparrow$
& SGF$\uparrow$
& IBC$\uparrow$
& ATA$\uparrow$
& PCA$\uparrow$
& WCS$\uparrow$
& MPA$\uparrow$
& RCA$\uparrow$
&
& \textbf{Score}$\uparrow$ \\
\midrule

\multicolumn{12}{c}{\emph{Open-source multimodal models}} \\
\midrule
Qwen3-VL-8B-Instruct~\citep{bai2025qwen3vl}
& 28.97 & 25.31 & 7.56 & 3.78
& 28.40 & 16.81 & 16.65
& 6.00 & 4.25 & 14.05 & 1.66 \\

DeepSeek-VL2-Small~\citep{wu2024deepseekvl2}
& 0.47 & 3.72 & 3.78 & \textemdash
& 5.89 & \textemdash & 1.58
& 3.00 & 7.00 & 3.16 & 0.41 \\

InternVL2.5-8B~\citep{chen2024internvl25}
& 8.88 & 17.57 & 10.50 & 3.36
& \cellcolor{secondcolor}\underline{29.91}
& 9.37 & 13.41
& 5.00 & 2.50 & 10.46 & 1.10 \\

Kimi-VL-A3B-Instruct~\citep{kimiteam2025kimivl}
& 0.93 & 5.85 & 3.78 & \textemdash
& \cellcolor{secondcolor}\underline{29.91}
& 18.84 & 17.96
& 7.00 & 8.00 & 10.79 & 0.88 \\

MiniCPM-V-2.6~\citep{yao2024minicpmv}
& 2.80 & 16.67 & 1.68 & \textemdash
& \cellcolor{secondcolor}\underline{29.91}
& 18.74 & 17.89
& \textemdash & 1.00 & 9.32 & 0.96 \\

LLaVA-OneVision-7B~\citep{li2024llavaonevision}
& \textemdash & 10.84 & \textemdash & \textemdash
& 17.07 & 3.29 & 6.66
& 1.00 & 2.75 & 4.53 & 0.61 \\

Molmo2-8B~\citep{clark2026molmo2}
& 26.17 & 17.86 & 23.53 & 11.76
& 24.92 & 6.28 & 13.62
& 7.00 & 11.50 & 14.67 & 1.48 \\

\midrule
\multicolumn{12}{c}{\emph{Proprietary multimodal models}} \\
\midrule
Claude-Sonnet-4.6~\citep{anthropic2026sonnet46}
& 13.64 & 22.06 & 22.92 & 12.50
& 28.79 & 18.27 & 16.21
& 10.00 & 16.25 & 17.33 & 1.60 \\

GPT-5.4-Mini~\citep{openai2026gpt54mini}
& \textemdash & 6.55 & 8.33 & \textemdash
& 20.45 & 8.65 & 13.02
& 10.00 & 16.25 & 10.30 & 1.25 \\

Qwen3-VL-Flash~\citep{bai2025qwen3vl}
& 18.18
& \cellcolor{bestcolor}\textbf{35.83}
& \cellcolor{secondcolor}\underline{31.25}
& 12.50
& 28.79
& 0.48
& 8.12
& 5.00
& 8.75
& 14.59
& 1.51 \\

Gemini-2.5-Flash~\citep{google2025gemini25}
& \cellcolor{secondcolor}\underline{38.64}
& 28.13
& 20.83
& \cellcolor{secondcolor}\underline{14.58}
& 27.27
& 10.10
& 11.98
& \cellcolor{secondcolor}\underline{15.00}
& \cellcolor{secondcolor}\underline{18.75}
& \cellcolor{secondcolor}\underline{19.62}
& \cellcolor{secondcolor}\underline{1.81} \\

\midrule
\multicolumn{12}{c}{\emph{GUI agent baselines}} \\
\midrule
AgentCPM-GUI~\citep{zhang2025agentcpmgui}
& \textemdash & 6.67 & \textemdash & \textemdash
& 29.61 & 18.94 & 17.83
& 1.00 & 6.00 & 9.10 & 0.55 \\

ScaleCUA-7B~\citep{liu2025scalecua}
& 17.29 & 17.48 & 4.62 & 2.10
& \cellcolor{secondcolor}\underline{29.91}
& \cellcolor{secondcolor}\underline{19.13}
& \cellcolor{secondcolor}\underline{18.02}
& 4.00 & 6.75 & 12.70 & 1.43 \\

UI-TARS-1.5-7B~\citep{qin2025uitars}
& 1.40 & 16.67 & 0.84 & \textemdash
& 24.92 & 13.04 & 13.59
& \textemdash & 7.25 & 8.51 & 0.98 \\

GUI-Owl-1.5-8B~\citep{xu2026mobileagent35}
& 10.75 & 22.26 & 12.61 & 2.10
& 28.70 & 18.36 & 17.21
& 12.00 & 17.00 & 15.95 & 1.53 \\

\midrule
\textbf{MAGAR(Ours)}
& \cellcolor{bestcolor}\textbf{39.25}
& \cellcolor{secondcolor}\underline{33.62}
& \cellcolor{bestcolor}\textbf{44.54}
& \cellcolor{bestcolor}\textbf{35.29}
& \cellcolor{bestcolor}\textbf{48.19}
& \cellcolor{bestcolor}\textbf{37.39}
& \cellcolor{bestcolor}\textbf{54.24}
& \cellcolor{bestcolor}\textbf{30.00}
& \cellcolor{bestcolor}\textbf{48.75}
& \cellcolor{bestcolor}\textbf{41.39}
& \cellcolor{bestcolor}\textbf{2.82} \\
\bottomrule
\end{tabular}
}
\end{table*}

The task-wise results clarify the complementary contributions of the two
intervention modules.
For Assembly Decision, Subgoal Fidelity rises from 42.86 to 44.54
and Intent Binding Consistency from 33.61 to 35.29. For Atomic Action,
Parameter Correctness Accuracy rises from 29.37 to 37.39 and Workflow
Consistency Score from 50.64 to 54.24. For Minimal Rework, Moved-Part
Accuracy rises from 24.00 to 30.00 and RCA from 45.50 to 48.75. These measurements connect the improvements to
structured intent, action parameterization, and recovery scope.

Supervised adaptation accounts for a substantial part of the overall
performance: the unadapted Qwen3-VL backbone scores 14.05, compared
with 37.50 after SFT. The intervention framework provides further
gains over this shared anchor. On the fixed judgment subset, MAGAR
receives an automatic score of 2.82, compared with 2.66 for Vanilla-SFT.

\subsection{Ablation Study}
\label{sec:ablation}
Table~\ref{tab:ablation} examines the contributions of the two
intervention modules using the same SFT anchor. Module~A improves
Assembly Decision and Minimal Rework, raising Overall from 37.50 to
39.63 while leaving Atomic Action unchanged. Module~B improves Atomic
Action prediction and reaches 39.26 Overall, with its largest
task-metric gain in parameter correctness. MAGAR applies the
corresponding module to each task and achieves 41.39 Overall. These
results support the framework's division between structured-field
correction and complete-action selection. The comparisons include both
candidate generation and candidate selection within each intervention
module; they do not constitute a separate ablation of the common
selection gate itself.

\begin{table*}[!htbp]
\centering
\footnotesize
\caption{
Ablation of the two intervention modules. The best results are
highlighted in bold.
}
\label{tab:ablation}
\setlength{\tabcolsep}{2.5pt}
\renewcommand{\arraystretch}{0.95}
\resizebox{\textwidth}{!}{
\begin{tabular}{l|cccc|ccc|cc|c|c}
\toprule
\multirow{2}{*}{\textbf{Model}}
& \multicolumn{4}{c|}{\textbf{Assembly Decision}}
& \multicolumn{3}{c|}{\textbf{Atomic Action}}
& \multicolumn{2}{c|}{\textbf{Minimal Rework}}
& \multirow{2}{*}{\textbf{Overall}$\uparrow$}
& \textbf{LLM-as-judge} \\
\cmidrule(lr){2-5}
\cmidrule(lr){6-8}
\cmidrule(lr){9-10}
& OPG$\uparrow$
& OBA$\uparrow$
& SGF$\uparrow$
& IBC$\uparrow$
& ATA$\uparrow$
& PCA$\uparrow$
& WCS$\uparrow$
& MPA$\uparrow$
& RCA$\uparrow$
&
& \textbf{Score}$\uparrow$ \\
\midrule
Vanilla-SFT
& 36.45 & 32.76 & 42.86 & 33.61
& 43.96 & 29.37 & 50.64
& 24.00 & 45.50 & 37.50 & 2.66 \\
SFT + Module A
& \cellcolor{bestcolor}\textbf{39.25}
& \cellcolor{bestcolor}\textbf{33.62}
& \cellcolor{bestcolor}\textbf{44.54}
& \cellcolor{bestcolor}\textbf{35.29}
& 43.96 & 29.37 & 50.64
& \cellcolor{bestcolor}\textbf{30.00}
& \cellcolor{bestcolor}\textbf{48.75}
& 39.63 & 2.67 \\
SFT + Module B
& 36.45 & 32.76 & 42.86 & 33.61
& \cellcolor{bestcolor}\textbf{48.19}
& \cellcolor{bestcolor}\textbf{37.39}
& \cellcolor{bestcolor}\textbf{54.24}
& 24.00 & 45.50 & 39.26 & 2.79 \\
\midrule
\textbf{MAGAR(Ours)}
& \cellcolor{bestcolor}\textbf{39.25}
& \cellcolor{bestcolor}\textbf{33.62}
& \cellcolor{bestcolor}\textbf{44.54}
& \cellcolor{bestcolor}\textbf{35.29}
& \cellcolor{bestcolor}\textbf{48.19}
& \cellcolor{bestcolor}\textbf{37.39}
& \cellcolor{bestcolor}\textbf{54.24}
& \cellcolor{bestcolor}\textbf{30.00}
& \cellcolor{bestcolor}\textbf{48.75}
& \cellcolor{bestcolor}\textbf{41.39}
& \cellcolor{bestcolor}\textbf{2.82} \\
\bottomrule
\end{tabular}
}
\end{table*}
\FloatBarrier
\subsection{Case Study and Error Analysis}
\label{sec:case_study}
\begin{figure}[t]
\centering
\includegraphics[width=\linewidth]{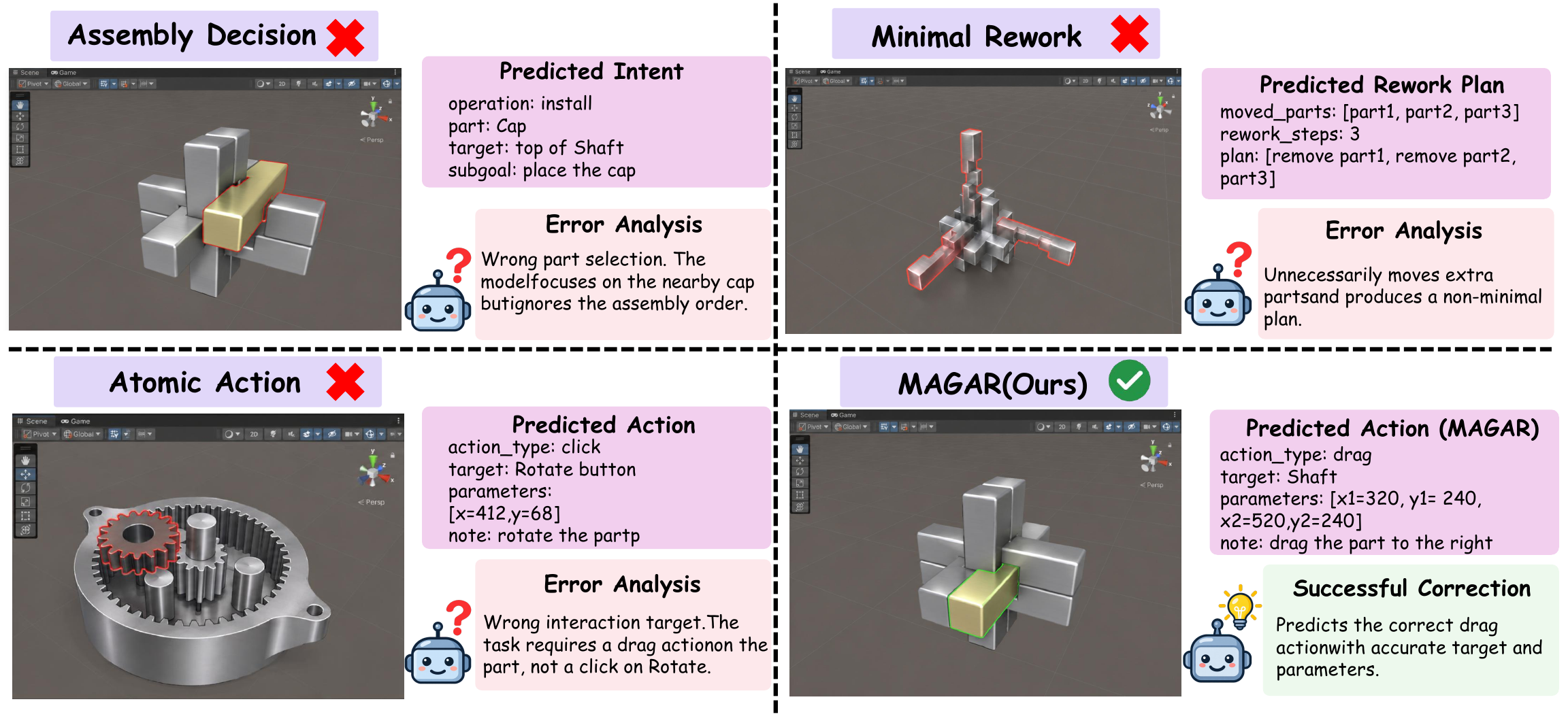}
\caption{Representative MAGAR corrections across three task families.}
\label{fig:case_study}
\end{figure}
Figure~\ref{fig:case_study} presents representative cases from the three
MAGAR task families, showing how task-specific intervention modifies
Vanilla-SFT predictions. For Assembly Decision, the intervention
revises inconsistent operation, involved-part, or subgoal predictions
while preserving reliable fields. For Atomic Action, MAGAR corrects
action grounding and GUI-action parameters, such as the interaction
target and its coordinates. For Minimal Rework, it improves the
identification of parts that must be moved and the minimum rework count.
These cases complement Table~\ref{tab:main_results}: Module~A addresses
structural assembly and rework reasoning, whereas Module~B targets
precise native GUI execution.

\subsection{Consistency with Human Judgments}
\label{sec:human_judgment}

\noindent
\begin{minipage}[t]{0.43\textwidth}
\vspace{0pt}

Figure~\ref{fig:judgment_consistency} shows how automatic scores
relate to human ratings across the evaluated models. Models with lower
LLM-as-judge scores generally receive lower human ratings, while
MAGAR lies in the high-scoring region on both axes, with an automatic
score of 2.82 and a mean human score of 2.77. The strong correlation
($r=0.9852$) indicates that the two evaluations track similar
model-level performance trends. The automatic scores therefore provide
a complementary signal for comparing relative performance across
models. This comparison concerns aggregated model scores rather than
agreement on individual predictions.
\end{minipage}
\hfill
\begin{minipage}[t]{0.54\textwidth}
\vspace{0pt}
\centering

\includegraphics[width=\linewidth]{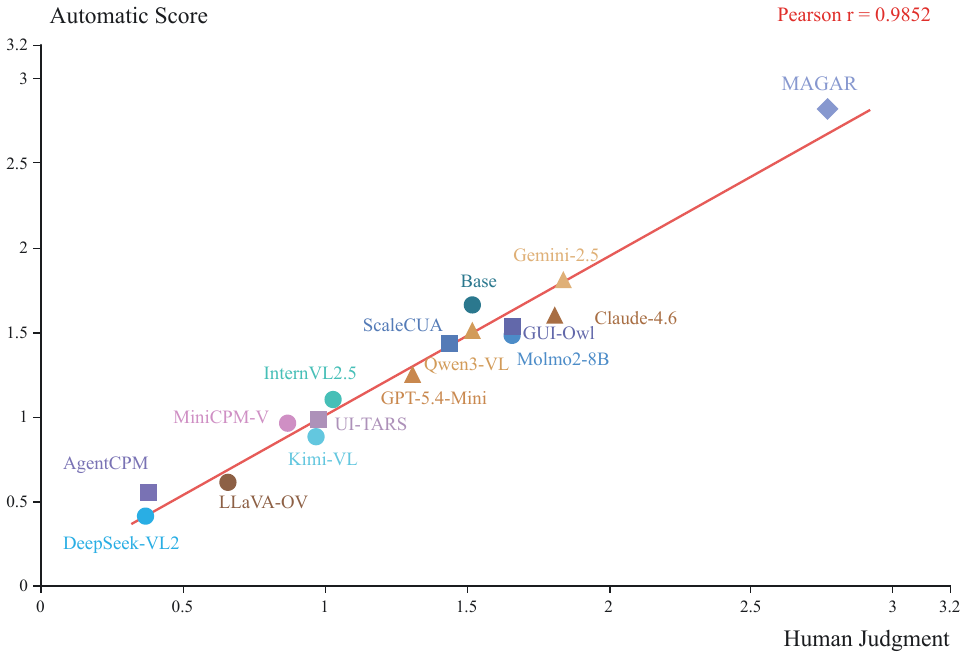}

\captionof{figure}{
Automatic evaluation score vs. human rating.
}
\label{fig:judgment_consistency}

\end{minipage}


\section{Conclusion and Limitations}
\label{sec:conclusion}
GUI-based mechanical assembly requires predictions that connect part relationships, native GUI actions, and recovery scope to the current assembly state. We introduced \textbf{MAGAR-Bench}, a benchmark that separates Assembly Decision, Atomic Action, and Minimal Rework, together with \textbf{MAGAR}, a selective correction framework built on a frozen SFT policy. MAGAR-Bench defines rework in terms of distinct moved parts, making preservation of the existing assembly an explicit recovery objective. MAGAR matches correction granularity to output dependencies through structured assembly and rework intervention in Module~A and complete GUI action intervention in Module~B. On recorded states from held-out assembly families, MAGAR improves the task balanced score from 37.50 to 41.39, with gains across all three tasks. Among 209 interventions, 70 correct an error, and 22 introduce one, highlighting both the benefit and the remaining cost of selective correction. Future work should extend this setting to long-horizon online assembly, globally consistent recovery planning, and broader CAD environments.


\clearpage
\section*{AI Use Statement}

Generative AI assistance was used for literature processing, dataset
construction, and auxiliary LLM-as-Judge evaluation.
All generated records, annotations, and evaluation outputs were reviewed under
the documented audit procedure.
The authors are responsible for the final dataset, experimental results,
analyses, and claims.

\section*{Reproducibility Statement}

Appendix~A reports the aggregate and radar-chart results. Appendix~B
documents MAGAR-Bench construction, dataset composition, and data splits.
Appendix~C provides the training configuration, decoding settings, and
evaluation metrics. Appendix~D contains the prompt templates, Appendix~E
presents the error analyses, and Appendix~F describes the human-evaluation
protocol and model-level agreement. All reported results use the fixed data
splits and evaluation protocols described in these appendices.

\bibliographystyle{iclr2027_conference}
\bibliography{iclr2027_conference}


\appendix
\clearpage
\section{Performance Results with Radar Chart Analysis}
\label{app:radar_results}

\begin{figure}[ht]
    \centering
    \includegraphics[width=0.9\linewidth]{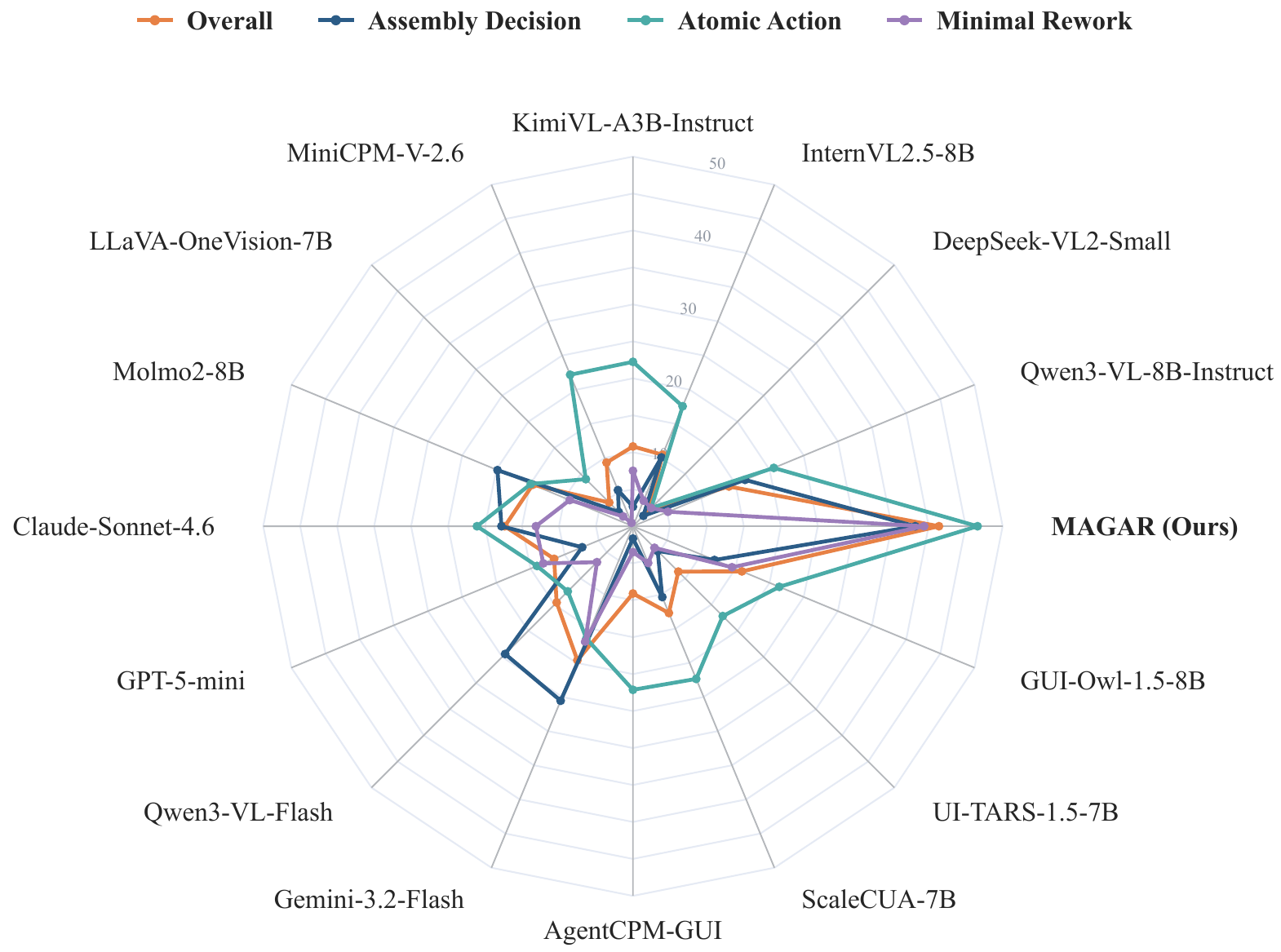}
    \caption{
    The curves report Overall, Assembly Decision, Atomic Action, and the
task-balanced Minimal Rework score, computed as the mean of MPA and RCA.
    }
    \vspace{-3mm}
    \label{fig:radar_chart}
\end{figure}

To provide a task-level view of model behavior, we compare MAGAR with
fifteen baselines from three categories: open-source multimodal models,
proprietary multimodal models, and GUI-specialized agents. Figure~\ref{fig:radar_chart}
summarizes four scores: Assembly Decision, Atomic Action, Minimal Rework,
and Overall. The Assembly Decision score is the mean of OPG, OBA, SGF,
and IBC. The Atomic Action score is the mean of ATA, PCA, and WCS. The
Minimal Rework score is the mean of MPA and RCA. Overall is computed using
the task-balanced aggregation defined in Appendix~\ref{app:metrics}.

The radar chart reveals substantial performance imbalance among existing
models. Most baseline systems produce uneven profiles across the three
task families. Several proprietary and GUI-specialized models obtain
relatively stronger scores on Atomic Action, but their performance
decreases on Assembly Decision and Minimal Rework. This pattern suggests
that producing a locally executable GUI action is easier than maintaining
consistency between the intended operation, the involved parts, the
current assembly state, and the required recovery scope. Open-source
models generally remain closer to the center of the chart, indicating
that their errors are distributed across structural decisions, action
parameters, and rework prediction.

MAGAR forms the largest and most balanced profile in the radar chart. It
achieves an Assembly Decision score of 38.18 and an Atomic Action score
of 46.61. The Minimal Rework score, computed as the mean of MPA and RCA,
is 39.38, resulting in an Overall score of 41.39. The higher Atomic Action
score reflects the benefit of task-specific action intervention and
complete primitive--parameter selection. The Assembly Decision score
indicates that structured-field intervention improves operation, part-set,
subgoal, and joint-intent prediction. The Minimal Rework score shows that
the intervention mechanism improves moved-part and count prediction while
retaining the reliable recovery fields of the SFT anchor.

The comparison also shows that strong performance on one task family does
not guarantee consistent performance across the benchmark. Models with
moderate action-grounding scores can still fail on assembly-level
constraints or recovery-scope decisions. MAGAR reduces this task-family
imbalance through separate candidate generators and selectors for
structured assembly outputs and complete GUI actions. Its larger and more
uniform radar profile supports the complementary contributions of the two
task-specific modules to assembly-consistent multimodal GUI prediction.

\section{MAGAR-Bench Details}
\label{app:dataset}

MAGAR-Bench is constructed from recorded Unity assembly episodes and
retains synchronized visual observations, public state, interaction
history, and structured supervision. Its assembly-family-disjoint split
assigns 20 families to training and six held-out families to Test, so an
assembly family appears in only one split. Table~\ref{tab:app_capability_coverage}
positions the benchmark relative to representative GUI and assembly
benchmarks in terms of their explicit evaluation targets.

\begin{table*}[ht]
\centering
\small
\setlength{\tabcolsep}{6pt}
\renewcommand{\arraystretch}{1.15}
\caption{Capability coverage of related benchmarks. A check mark denotes
explicit coverage, and X denotes that the capability is not a dedicated
evaluation target.}
\label{tab:app_capability_coverage}
\begin{tabular}{lccccc}
\toprule
\textbf{Benchmark / Work}
& \makecell{\textbf{GUI}}
& \makecell{\textbf{Config.}\\\textbf{Target}}
& \makecell{\textbf{3D}\\\textbf{Assembly}}
& \makecell{\textbf{Error}\\\textbf{Repair}}
& \makecell{\textbf{Min.}\\\textbf{Moved Parts}} \\
\midrule
VisualWebArena~\citep{koh2024visualwebarena}
& $\checkmark$ & X & X & X & X \\
OSWorld~\citep{xie2024osworld}
& $\checkmark$ & X & X & X & X \\
AndroidWorld~\citep{rawles2025androidworld}
& $\checkmark$ & X & X & X & X \\
PAGE Bench / PAGER~\citep{wei2026pager}
& $\checkmark$ & $\checkmark$ & X & X & X \\
LEGO-ECA / SCANet~\citep{wan2024scanet}
& X & $\checkmark$ & $\checkmark$ & $\checkmark$ & X \\
FurnitureBench~\citep{heo2025furniturebench}
& X & $\checkmark$ & $\checkmark$ & X & X \\
\midrule
\rowcolor{blue!8}
\textbf{MAGAR-Bench}
& $\checkmark$ & $\checkmark$ & $\checkmark$
& $\checkmark$ & $\checkmark$ \\
\bottomrule
\end{tabular}
\end{table*}

\FloatBarrier


\section{Experimental Details}
\label{app:experiments}

\subsection{Training and Decoding Configuration}
\label{app:training_configuration}

MAGAR uses \texttt{Qwen3-VL-8B-Instruct} as its backbone.
Table~\ref{tab:app_training_configuration} summarizes the supervised
adaptation and decoding configuration. The visual encoder and multimodal
aligner remain frozen during LoRA adaptation, after which the SFT policy
serves as the frozen anchor for the two intervention modules.

\begin{table}[t]
\centering
\small
\caption{Backbone adaptation and decoding configuration.}
\label{tab:app_training_configuration}
\begin{tabular}{ll}
\toprule
Setting & Value \\
\midrule
Backbone & \texttt{Qwen3-VL-8B-Instruct} \\
Adaptation & LoRA \\
LoRA rank / scaling factor & 8 / 32 \\
LoRA dropout & 0.05 \\
Frozen components & Visual encoder and multimodal aligner \\
Optimizer & AdamW \\
Learning rate & $1\times10^{-4}$ \\
Training epochs & 2 \\
Effective batch size & 16 \\
Maximum sequence length & 8,192 tokens \\
Numerical precision & bfloat16 \\
Hardware & Two NVIDIA A100-SXM4-40GB GPUs \\
Decoding & Deterministic greedy decoding \\
Maximum generated tokens & 256 \\
\bottomrule
\end{tabular}
\end{table}

\subsection{Minimal-Rework Ground-Truth Construction}
\label{app:minimal_rework_ground_truth}

For each Minimal-Rework sample, the reference is selected from the
validated recovery candidates associated with the failed and target
states. Candidates are compared by the number of distinct moved parts,
and the selected reference defines the minimum moved-part count within
that validated candidate set. This benchmark does not claim unrestricted
global optimality over all possible GUI action sequences. Recovery route
and corrective action identifiers are retained as auxiliary fields; their
matching does not verify execution success.

\subsection{Evaluation Metrics}
\label{app:metrics}

We evaluate every prediction with the frozen component-resolved evaluator.
The nine primary metrics are reported as percentages. Invalid or
unparseable predictions receive zero credit for the affected components and
remain in the corresponding denominator. A dash in a result table denotes an
observed score of zero (rather than a missing observation) and is retained as
zero when computing Overall.

\subsubsection{Assembly-Decision Metrics}

The evaluator scores four complementary properties of the predicted assembly
decision. An invalid structured decision receives no field-level credit.
\textbf{Operation--Part Grounding (OPG)} is the fraction of samples with a
nonempty reference part set for which both the operation and the involved-part
set are exact; only these applicable samples enter its denominator.
\textbf{Operation-Balanced Accuracy (OBA)} is the unweighted mean of
per-operation recall over operation classes represented in the reference set.
\textbf{Subgoal Fidelity (SGF)} is the fraction of samples whose next
subgoal is exact after Unicode-text normalization. \textbf{Intent Binding
Consistency (IBC)} is the fraction for which the operation, involved-part set,
and next subgoal are jointly correct. Part IDs are compared as normalized
sets, so their order does not affect any of these judgments; the high-level
task ID is not included.

\subsubsection{Atomic-Action Metrics}

\textbf{Action-Type Accuracy (ATA)} is the fraction of samples whose native
action type matches the reference. The evaluator resolves documented
primitive-key aliases before comparing action types, without relaxing
parameter matching. \textbf{Parameter Correctness Accuracy (PCA)} is
micro-averaged over all required action-specific parameter components, and a
component receives credit only when its action type is correct. For a click,
the button must match and the predicted screen point must lie in the
reference hit range. For a drag, the button and both endpoint hit ranges must
match, and the predicted duration must be within the larger of $50$ ms and
$10\%$ of the reference duration $d$. Text is matched exactly, while keys
and modifier sets are compared case-insensitively.

\textbf{Workflow Consistency Score (WCS)} first averages the correct type and
parameter components at each step, then averages steps within each recorded
high-level task, and finally assigns equal weight to the high-level tasks.
Thus, WCS measures component coverage rather than the fraction of workflows
whose complete action sequence is exact.

\subsubsection{Minimal-Rework Metrics}

For the $N_{\rm MR}=100$ Test rework samples, each reference and
prediction contains a moved-part set $P_m$, a moved-part count $n_m$,
a recovery route ID $R$, and an ordered corrective action-ID sequence
$A$. Part IDs are compared as normalized sets, and the moved-part count
is checked against the cardinality of the predicted set. The recovery
route ID and corrective action sequence describe the associated recovery
process and contribute to RCA through identifier matching; this
comparison does not execute the recovery.

Let $b_P$, $b_n$, $b_R$, and $b_A$ denote binary correctness indicators
for the moved-part set, moved-part count, recovery route ID, and
corrective action-ID sequence, respectively.
\begin{equation}
\begin{aligned}
\mathrm{MPA}
&=
\frac{1}{N_{\rm MR}}
\sum_{i=1}^{N_{\rm MR}}
\mathbf{1}\!\left[
b_{P,i}\land b_{n,i}
\land |\widehat{P}_{m,i}|=\widehat{n}_{m,i}
\right],\\[4pt]
\mathrm{RCA}
&=
\frac{1}{N_{\rm MR}}
\sum_{i=1}^{N_{\rm MR}}
\frac{
b_{P,i}+b_{n,i}+b_{R,i}+b_{A,i}
}{4}.
\end{aligned}
\label{eq:app_rework_metrics}
\end{equation}
\textbf{Moved-Part Accuracy (MPA)} requires the moved-part set and its
count to be jointly correct and internally consistent. \textbf{RCA}
is a four-field consistency score that averages correctness of the
moved-part set, count, recovery route ID, and corrective action-ID
sequence. The recovery-objective string and shortest-path claim are
diagnostic checks and do not contribute to MPA or RCA.

\subsubsection{Overall Score and LLM-as-Judge}

Overall first averages the automatic metrics within each task family and then
assigns equal weight to the three family-level scores:
\begin{equation}
\mathrm{Overall}
=\frac{1}{3}\left(
\frac{\mathrm{OPG}+\mathrm{OBA}+\mathrm{SGF}+\mathrm{IBC}}{4}
+\frac{\mathrm{ATA}+\mathrm{PCA}+\mathrm{WCS}}{3}
+\frac{\mathrm{MPA}+\mathrm{RCA}}{2}
\right).
\label{eq:overall_metric}
\end{equation}

The LLM-as-judge score is reported separately and is not included in Overall.
The automatic judge is \texttt{qwen3.8-flash}, run with temperature $0$ and
no reasoning effort. LLM-as-judge and human evaluation use the same fixed
subset of 120 Test instances, obtained by stratified random sampling with
40 instances from each task family. The judge assigns a continuous score
from $0.00$ to $5.00$ to each model-output item and is used only as an
auxiliary comparison.
\subsection{Fine-Grained Performance}
\label{app:fine_grained_analysis}

\begin{figure}[ht]
\centering
\includegraphics[width=\linewidth]{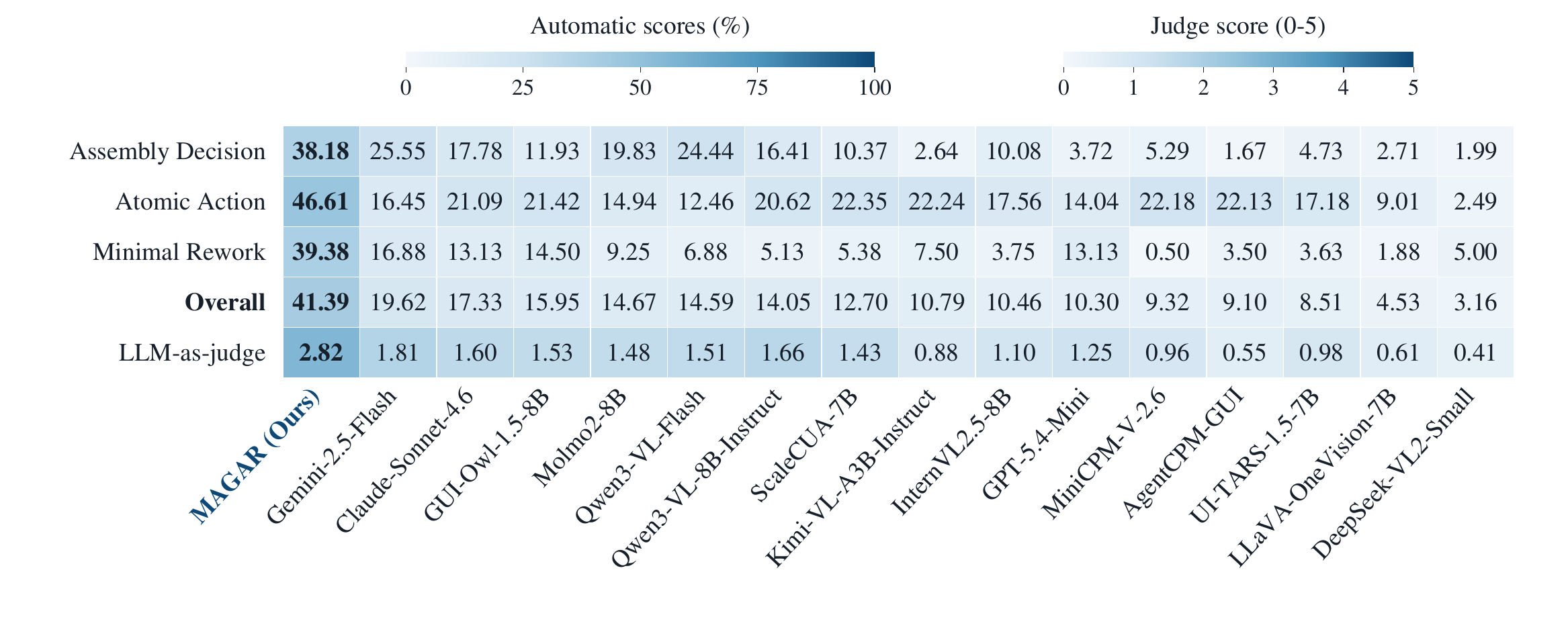}
\caption{Fine-grained performance across task subcategories. Rows denote
subcategories of Assembly Decision, Atomic Action, and Minimal Rework, and
columns denote evaluated models.}
\label{fig:fine_grained_performance_heatmap}
\end{figure}

Figure~\ref{fig:fine_grained_performance_heatmap} decomposes the aggregate
results into recorded subcategories. The Assembly Decision rows cover
operation selection, part identification, subgoal prediction, and joint
intent consistency. Atomic Action separates click, drag, keyboard, and text
interactions, while Minimal Rework is grouped by the number of parts that
must move.

\FloatBarrier

\section{Prompt Templates}
\label{app:prompt}

This appendix provides the task-specific prompt templates used by
MAGAR-Bench. Each template follows the corresponding dataset schema and
output contract. The reference assistant response is removed during
inference, while the input fields and formatting remain unchanged.

All three tasks use the same flat-output convention: the response is
exactly one JSON object, the required fields are fixed by the task
family, field names and value types must be preserved, and no wrapper
object is used. Explanations, Markdown fences, comments, and additional
fields are not allowed. The task family is supplied by the input prompt
and is not repeated as an output field.


\subsection{Assembly Decision Prompt}
\label{app:prompt_assembly}

\begin{promptbox}[Assembly Decision]

\sloppy
\setlength{\emergencystretch}{1em}

\textbf{Role.}

You are a multimodal assembly reasoning agent. Your task is to predict
the next valid assembly intent from the current Unity state, the target
assembly, the available part references, the public state, and the
interaction history.

\vspace{2pt}
\textbf{Task family.}

The task family is
\texttt{assembly\_decision}. The prediction describes a structured
assembly decision rather than a low-level GUI action.

The allowed values of \texttt{operation} are
\texttt{align}, \texttt{execute\_subtask}, \texttt{inspect},
\texttt{lock}, \texttt{place}, and \texttt{recover}. A spatial or
procedural phrase such as ``insert along axis'' belongs in
\texttt{next\_subgoal}, not in \texttt{operation}.

\vspace{2pt}
\textbf{Available input.}

The input contains:

\begin{itemize}
    \item the current Unity observation:
    \textless CURRENT\_UNITY\_IMAGE\textgreater;
    \item the target assembly image:
    \textless TARGET\_ASSEMBLY\_IMAGE\textgreater;
    \item candidate-part reference images:
    \textless PART\_REFERENCE\_IMAGES\textgreater;
    \item the public state JSON:
    \textless PUBLIC\_STATE\_JSON\textgreater;
    \item the summarized interaction history:
    \textless HISTORY\_SUMMARY\textgreater.
\end{itemize}

\vspace{2pt}
\textbf{Reasoning requirements.}

\begin{enumerate}
    \item Identify the next operation only from the current state, target
    state, public state, and interaction history.
    \item Select involved-part IDs only from the provided part references
    and public-state JSON.
    \item Preserve the association between the predicted operation and
    its involved parts.
    \item Predict the next subgoal as a concise continuation of the
    recorded task state.
    \item Do not output a GUI action, screen coordinate, route ID,
    corrective action, or private verifier field.
    \item Use exactly one allowed operation value and do not introduce a
    part ID or subgoal that is not supported by the input.
\end{enumerate}

\vspace{2pt}
\textbf{Output requirements.}

Return exactly one JSON object with the following three fields, in the
order shown:

\begin{itemize}
    \item \texttt{operation};
    \item \texttt{involved\_part\_ids};
    \item \texttt{next\_subgoal}.
\end{itemize}

Use the exact identifiers and text conventions provided by the dataset.
Do not change the field names, add extra fields, or include an explanation.

\vspace{2pt}
\textbf{Output format.}

{\scriptsize
\begin{tabular}{@{}l@{}}
\texttt{\{} \\
\hspace*{1em}\texttt{"operation": "<operation>",} \\
\hspace*{1em}\texttt{"involved\_part\_ids": ["<part\_id>"],} \\
\hspace*{1em}\texttt{"next\_subgoal": "<next\_subgoal>"} \\
\texttt{\}}
\end{tabular}
}

\vspace{2pt}
\textbf{Final-output rule.}

Return the JSON object only. Do not wrap it in a Markdown code block.

\end{promptbox}

\captionof{figure}{Prompt template for the Assembly Decision task.}
\label{fig:app_assembly_prompt}


\subsection{Atomic Action Prompt}
\label{app:prompt_atomic}

\begin{promptbox}[Atomic Action]

\sloppy
\setlength{\emergencystretch}{1em}

\textbf{Role.}

You are a GUI action execution agent. Your task is to predict exactly
one executable native GUI action for the current Unity interface.

\vspace{2pt}
\textbf{Task family.}

The task family is
\texttt{atomic\_action}. The output must contain one complete primitive
and its native parameters.

\vspace{2pt}
\textbf{Available input.}

The input contains:

\begin{itemize}
    \item the current Unity frame:
    \textless CURRENT\_UNITY\_FRAME\textgreater;
    \item the current assembly subgoal:
    \textless CURRENT\_SUBGOAL\textgreater;
    \item the public GUI state:
    \textless PUBLIC\_STATE\_JSON\textgreater;
    \item the summarized interaction history:
    \textless HISTORY\_SUMMARY\textgreater.
\end{itemize}

\vspace{2pt}
\textbf{Reasoning requirements.}

\begin{enumerate}
    \item Select exactly one native action type from
    \texttt{click}, \texttt{drag}, \texttt{type\_text},
    \texttt{keypress}, and \texttt{hotkey}.
    \item Predict parameters in the native format expected by the Unity
    interface.
    \item Use screen coordinates only when they are supported by the
    current frame and public GUI state.
    \item Preserve the pairing between the action type and its
    parameters.
    \item Do not output a semantic part ID, normalized coordinate,
    bounding box, verifier field, future action, or action sequence.
    \item Do not invent a coordinate, button, key, modifier, text string,
    or duration that is not supported by the input.
\end{enumerate}

\vspace{2pt}
\textbf{Allowed action contracts.}

\begin{itemize}
    \item \texttt{click}: \texttt{type}, \texttt{button},
    \texttt{screen\_point\_px};
    \item \texttt{drag}: \texttt{type}, \texttt{button},
    \texttt{start\_screen\_point\_px},
    \texttt{end\_screen\_point\_px}, \texttt{duration\_ms};
    \item \texttt{type\_text}: \texttt{type}, \texttt{text};
    \item \texttt{keypress}: \texttt{type}, \texttt{key};
    \item \texttt{hotkey}: \texttt{type}, \texttt{key},
    \texttt{modifiers}.
\end{itemize}

\vspace{2pt}
\textbf{Output requirements.}

Return exactly one JSON object for one action. The field \texttt{type}
is the action discriminator and must be one of the five allowed native
operations. Include exactly the parameter fields associated with that
operation, using the native parameter names specified above. Do not
include reasoning, a task-family field, or additional fields.

\vspace{2pt}
\textbf{Output format for a click.}

{\scriptsize
\begin{tabular}{@{}l@{}}
\texttt{\{} \\
\hspace*{1em}\texttt{"type": "click",} \\
\hspace*{1em}\texttt{"button": "<button>",} \\
\hspace*{1em}\texttt{"screen\_point\_px": [<x\_px>, <y\_px>]} \\
\texttt{\}}
\end{tabular}
}

\vspace{2pt}
\textbf{Output format for a drag.}

{\scriptsize
\begin{tabular}{@{}l@{}}
\texttt{\{} \\
\hspace*{1em}\texttt{"type": "drag",} \\
\hspace*{1em}\texttt{"button": "<button>",} \\
\hspace*{1em}\texttt{"start\_screen\_point\_px": [<x\_1>, <y\_1>],} \\
\hspace*{1em}\texttt{"end\_screen\_point\_px": [<x\_2>, <y\_2>],} \\
\hspace*{1em}\texttt{"duration\_ms": <duration>} \\
\texttt{\}}
\end{tabular}
}

\vspace{2pt}
\textbf{Final-output rule.}

Return the JSON object only. Do not wrap it in a Markdown code block.

\end{promptbox}

\captionof{figure}{Prompt template for the Atomic Action task.}
\label{fig:app_atomic_prompt}


\subsection{Minimal Rework Prompt}
\label{app:prompt_rework}

\begin{promptbox}[Minimal Rework]

\sloppy
\setlength{\emergencystretch}{1em}

\textbf{Task.}

Given a failed assembly state, a target state, candidate-part references,
public recovery state, and interaction history, predict the recovery
information while minimizing the number of distinct moved parts. The task
family is \texttt{minimal\_rework}.

\vspace{2pt}
\textbf{Constraints.}

Use only the provided part, route, and action IDs. Return distinct
\texttt{moved\_part\_ids}, set
\texttt{min\_moved\_part\_count} to their cardinality, and preserve the
order of \texttt{action\_ids} for the selected route. The objective is
the minimum distinct moved-part objective within the provided recovery
candidates. The shortest-path field is diagnostic and should be
\texttt{true} only when explicitly supported by the input.

\vspace{2pt}
\textbf{Input.}

\texttt{failed\_assembly}, \texttt{target\_assembly},
\texttt{part\_references};\\[-1pt]
\texttt{public\_recovery\_state}, \texttt{history}.

\vspace{2pt}
\textbf{Output.}

Return exactly one JSON object with these six fields and no explanation,
Markdown fence, or additional field:

\vspace{2pt}
\textbf{Output format.}

{\scriptsize
\begin{tabular}{@{}l@{}}
\texttt{\{} \\
\hspace*{1em}\texttt{"objective": "minimum\_distinct\_moved\_parts",} \\
\hspace*{1em}\texttt{"selected\_route\_id": "<route\_id>",} \\
\hspace*{1em}\texttt{"moved\_part\_ids": ["<part\_id>"],} \\
\hspace*{1em}\texttt{"min\_moved\_part\_count": 1,} \\
\hspace*{1em}\texttt{"action\_ids": ["<action\_id>"],} \\
\hspace*{1em}\texttt{"full\_shortest\_path\_claim": false} \\
\texttt{\}}
\end{tabular}
}

\vspace{2pt}
\textbf{Final-output rule.} Return the JSON object only.

\end{promptbox}

\captionof{figure}{Prompt template for the Minimal Rework task.}
\label{fig:app_rework_prompt}

\FloatBarrier


\section{Error Analysis}
\label{app:error_analysis}

\subsection{Intervention Outcomes}
\label{app:intervention_outcomes}

Table~\ref{tab:app_intervention_accounting} reports how often the final
prediction differs from the SFT anchor. Across the 209 interventions, 70
change an incorrect prediction to a correct one, 22 change a correct
prediction to an incorrect one, and 117 leave strict correctness unchanged.
The last group includes correct-to-correct and incorrect-to-incorrect changes
that may still alter partial-credit metrics. These counts use a strict
case-level criterion and therefore should not be interpreted as the
task-balanced Overall score in Eq.~\ref{eq:overall_metric}.

\begin{table}[ht]
\centering
\small
\caption{Intervention frequency by task family. Rates use the corresponding
Test-set sizes.}
\label{tab:app_intervention_accounting}
\begin{tabular}{lrrr}
\toprule
Task family & Test samples & Interventions & Rate (\%) \\
\midrule
Assembly Decision & 238 & 14 & 5.88 \\
Atomic Action & 662 & 184 & 27.79 \\
Minimal Rework & 100 & 11 & 11.00 \\
\midrule
All tasks & 1,000 & 209 & 20.90 \\
\bottomrule
\end{tabular}
\end{table}

\subsection{Task-Specific Failure Modes}
\label{app:task_specific_errors}

The three tasks expose different ways an intervention can correct an anchor
prediction or introduce a new error. The intervention counts above do not
separately quantify these failure modes.

\paragraph{Assembly Decision.}
Corrections to the operation, involved-part set, or next subgoal can improve
individual fields without necessarily making the complete assembly intent
consistent. Conversely, changing a reliable field can turn an otherwise
correct anchor into an incorrect prediction.

\paragraph{Atomic Action.}
An action can have the correct type but incorrect parameters, such as a
misplaced click point or drag endpoint. Selecting a complete action keeps
the type and parameters together, but does not guarantee that either is
correctly grounded in the current GUI state.

\paragraph{Minimal Rework.}
The moved-part set can omit required parts or include unnecessary ones, and
the predicted count can disagree with the selected set. MPA evaluates the
moved-part set and count jointly, while RCA additionally checks recovery
route and corrective action identifiers. A higher RCA value still does not
establish that the retained route and action sequence form a valid recovery.

\FloatBarrier

\section{Human Evaluation}
\label{app:human}

\subsection{Protocol and Aggregation}
\label{app:human_protocol}

Human ratings and automatic LLM-as-judge scores are compared on the same
stratified-random subset of 120 Test instances, with 40 instances from each
task family. Each point in Figure~\ref{fig:judgment_consistency} represents
the aggregate automatic and human scores of one of the 16 evaluated models.
The comparison is therefore performed at the model level rather than at the
level of individual predictions.

The automatic scores are produced by \texttt{qwen3.8-flash} with temperature
$0$ and no reasoning effort. Human evaluation was conducted by three experts
with backgrounds in GUI interaction, mechanical assembly, or multimodal
models. All three experts independently rated every evaluated model output.
Annotators saw the same model outputs and task context used by the LLM judge,
applied the same scoring rubric, and were blinded to model identities.

For each model output, we first averaged the three expert scores with equal
weight. We then averaged these per-output scores over the 120 evaluation
instances to obtain the model-level human score. Because the subset contains
40 instances from each task family, this aggregation assigns equal weight to
the three task families.

\subsection{Model-Level Agreement}
\label{app:human_agreement}

Let $J_m$ and $H_m$ denote the aggregate automatic and human scores for
model $m$, respectively. For $M=16$ evaluated models, Pearson correlation is
computed as
\begin{equation}
r=
\frac{
\sum_{m=1}^{M}(J_m-\overline{J})(H_m-\overline{H})
}{
\sqrt{\sum_{m=1}^{M}(J_m-\overline{J})^2}
\sqrt{\sum_{m=1}^{M}(H_m-\overline{H})^2}
},
\label{eq:app_human_pearson}
\end{equation}
where
\begin{equation}
\overline{J}
=
\frac{1}{M}\sum_{m=1}^{M}J_m,
\qquad
\overline{H}
=
\frac{1}{M}\sum_{m=1}^{M}H_m.
\end{equation}

The reported model-level correlation is $r=0.9852$.
MAGAR obtains an automatic score of 2.82 and a mean human score
of 2.77. The correlation indicates that the two evaluations track similar
performance trends across models; it does not establish sample-level
agreement or agreement among individual annotators.

\end{document}

%% file: math_commands.tex
\usepackage{amsmath,amsfonts,bm}

\def\eqref#1{equation~\ref{#1}}

\def\1{\bm{1}}

\DeclareMathAlphabet{\mathsfit}{\encodingdefault}{\sfdefault}{m}{sl}
\SetMathAlphabet{\mathsfit}{bold}{\encodingdefault}{\sfdefault}{bx}{n}

